\documentclass[10pt, a4paper]{article}

\usepackage{lrec-coling2024} 

\usepackage{listings}
\usepackage{xcolor}
\usepackage{booktabs}
\usepackage{algorithm} 
\usepackage{algpseudocode}
\usepackage{amsmath}
\usepackage{pgfplots}
\usepackage{tabularx}
\usepackage{graphicx}
\usepackage[inline]{enumitem}
\usepackage{arydshln}
\usepackage{enumitem} 
\usepackage{ragged2e} 
\usepackage{hyperref}

\definecolor{codegreen}{rgb}{0,0.6,0}
\definecolor{codegray}{rgb}{0.5,0.5,0.5}

\usepackage{tcolorbox}
\tcbuselibrary{skins}

\definecolor{myblue}{RGB}{0,163,243}
\definecolor{lightblue}{RGB}{218,234,255}

\newtcolorbox{mybox}[2][]{%
  enhanced,
  colback=white,
  colframe=myblue,
  coltitle=myblue,
  sharp corners,
  boxrule=1pt,
  #1
}

\pgfmathsetseed{\number\pdfrandomseed} 

\pgfplotsset{compat=1.18}

\lstdefinestyle{regex}{
    basicstyle=\ttfamily\footnotesize,
    commentstyle=\color{codegreen},
    keywordstyle=\color{magenta},
    numberstyle=\tiny\color{codegray},
    stringstyle=\color{codegreen},
    breakatwhitespace=false,         
    breaklines=true,                 
    captionpos=b,                    
    keepspaces=true,                 
    showspaces=false,                
    showstringspaces=false,
    showtabs=false,                  
    tabsize=2,
    morecomment=[l]{\#}
}

\title{
  \vspace*{.5\baselineskip}
  \normalfont{
    \vspace*{.5\baselineskip}
    \textbf{
      ELLEN:
      E{\bf {\em xtremely}}
      L{\bf {\em ightly}}
      {\bf {\em Supervised}}
      L{\bf {\em earning}}
      {\bf {\em For}}
      E{\bf {\em fficient}}
      N{\bf {\em amed}}
      {\bf {\em Entity}}
      {\bf {\em Recognition}}
    }
  }
}

\name{Haris Riaz, Razvan-Gabriel Dumitru, Mihai Surdeanu} 

\address{University of Arizona \\
         Tucson, AZ, USA\\
         \{hriaz, rdumitru, msurdeanu\}@arizona.edu \\
         }

\abstract{
In this work, we revisit the problem of semi-supervised named entity recognition (NER) focusing on extremely light supervision, consisting of a lexicon containing only 10 examples per class.
We introduce ELLEN, a simple, fully modular, neuro-symbolic method that blends fine-tuned language models with linguistic rules. These rules include insights such as ``One Sense Per Discourse'', using a Masked Language Model as an unsupervised NER, leveraging part-of-speech tags to identify and eliminate unlabeled entities as false negatives, and other intuitions about classifier confidence scores in local and global context. ELLEN achieves very strong performance on the CoNLL-2003 dataset when using the minimal supervision from the lexicon above. It also outperforms most existing (and considerably more complex) semi-supervised NER methods under the same supervision settings commonly used in the literature (i.e., 5\% of the training data). Further, we evaluate our CoNLL-2003 model in a zero-shot scenario on WNUT-17 where we find that it outperforms GPT-3.5 and achieves comparable performance to GPT-4. In a zero-shot setting, ELLEN also achieves over 75\% of the performance of a strong, fully supervised model trained on gold data. Our code is publicly \href{https://github.com/hriaz17/ELLEN}{available}.
\newline\newline}

\begin{document}

\maketitleabstract

\section{Introduction}
Named entity recognition (NER), i.e., the task of identifying named (and sometimes numeric) entities such person and organization names, drugs, protein names, diseases, and dates, is one of the earliest formal natural language processing (NLP) tasks~\cite{grishman1996message}. NER remains critical to many real-world applications such as question answering and information extraction~\cite{yadav2019survey}. Despite the tremendous progress observed on the NER task in the past almost three decades, we argue that there are several practical limitations in the way this task is generally formalized, which impact our understanding of what methods perform best in practice. In particular:
{\flushleft {\bf (1)}} Current settings for the NER task require an amount of annotations that are unrealistic for many real-world applications. For example, a common setting for semi-supervised NER uses 5\% of the CoNLL-2003 corpus' (\citetlanguageresource{tjong-kim-sang-de-meulder-2003-introduction}) training data, or over 10K total tokens~\cite{chen-etal-2020-local,zheng-etal-2023-jointprop}. In our work \cite{vacareanu-etal-2024-active}, we have observed that NER annotations take approximately 3.2 seconds per token in practice. Thus, annotating the equivalent amount of data in a new domain would take approximately 9 person hours. This is unrealistic in many scenarios (e.g., intelligence, pandemic surveillance) that require the rapid development of custom models and where domain experts ``do not want to come willingly and do not come cheaply.''\footnote{IARPA program manager, personal communication}

{\flushleft {\bf (2)}} While recent directions that use in-context learning (ICL) for NER with autoregressive decoder-based large language models (LLMs) perform well \cite{chen2023learning}, they do not scale as well as encoder-based methods due to the decoder's high inference overhead; each generated token requires its own forward pass through the model. 

{\flushleft {\bf (3)}} 
Recent trends rely mostly on neural networks (NNs) to learn the NER task, ignoring linguistic hints such as ``one sense per discourse''~\cite{gale1992one} that might be present and are likely to be useful in lightly-supervised settings.

To remedy these limitations we propose an {\em extremely lightly supervised} scenario for NER, in which the only supervision comes in the form of a lexicon containing 10 examples per entity class. Importantly, the 10 examples are selected by a domain expert that does {\em not} have access to any dataset annotations. Further, we propose a simple NER approach for this scenario that is efficient and performs well despite the limited supervision. Our method uses an encoder-only inference strategy, but, at training time, it combines multiple strategies including language models and several linguistic heuristics. We call our method {\em {\em E}xtremely {\em L}ightly Supervised {\em L}earning for {\em E}fficient {\em N}amed Entity Recognition} (ELLEN)\footnote{\url{https://github.com/hriaz17/ELLEN}}.

Our main contributions are as follows:
{\flushleft {\bf (1)}} We demonstrate the effectiveness of combining language models with commonsense linguistic rules inspired by \cite{10.5555/1621829.1621837} and aggregated under a self-training, modular, neuro-symbolic architecture. Our approach is considerably simpler than other complex statistical methods for semi-supervised NER~\cite[inter alia]{nagesh-surdeanu-2018-keep,lakshmi-narayan-etal-2019-exploration, peng-etal-2019-distantly, zhou-etal-2022-distantly, chen2019variational, 46655, chen-etal-2020-local, zheng-etal-2023-jointprop}.

{\flushleft {\bf (2)}} Our approach includes a novel component called the Masked Language Modeling (MLM) Heuristic, which is a fully unsupervised NER method that achieves over 55\% precision on the CoNLL-2003 NER dataset. Further, this component complements other self training as well as linguistic heuristics in the semi-supervised setting.

{\flushleft {\bf (3)}} We evaluate our method on CoNLL-2003 (\citetlanguageresource{tjong-kim-sang-de-meulder-2003-introduction}) under three different degrees of supervision, and in a zero-shot setting on WNUT-17 (\citetlanguageresource{derczynski-etal-2017-results}). 
On CoNLL, under the proposed setting of extremely limited supervision, we show that our method achieves an F1 score of \textbf{76.87}\%. Further, when we increase the degree of supervision to match other methods which are state-of-the-art in the semi-supervised NER setting, we find that our method achieves comparable performance. We also show that our method continues to scale, even when using all of the data available for supervision. In a zero-shot evaluation on WNUT-17, we find our method to be comparable to LLMs such as GPT-3.5 \cite{openai_chatgpt} and GPT-4 \cite{OpenAI2023GPT4TR}, while also obtaining over 75\% of the performance of a fully supervised model trained on WNUT-17.




\section{Related Works}
Recently, Large Language Models (LLM's) have emerged as the dominant approach for a wide variety of NLP tasks, including Named Entity Recognition. \cite{wang2023gptner} and \cite{zhou2023universalner} show that LLM's consistenly achieve SOTA performance on many NER datasets. With In-Context Learning, LLM's have also proven to be very useful in the FewShot NER setting, as recently shown by \cite{ashok2023promptner}. However, LLM's typically have a high inference overhead \cite{narayanan2023cheaply} and may not always perform well in specialized or low-resource domains. Moreover, there are increasing concerns about data contamination. \cite{DBLP:journals/corr/abs-2308-08493} demonstrate that GPT-3.5 and GPT-4 have encountered test data with labels from widely-used NLP benchmark datasets during pre-training. \cite{sainz2023lm} claim that this is true for CoNLL-2003, one of our evaluation datasets.


Focusing on Semi-Supervised NER, and not FewShot NER, current state-of-the-art methods include JointProp \cite{zheng-etal-2023-jointprop}, which is a multi-task learning framework that jointly tries to solve relation classification and NER using a heterogeneous graph structure. Semi-LADA \cite{chen-etal-2020-local} adapts the mixup data augmentation technique to sequence labeling, and then trains on linearly interpolated pairs of samples. Both of these methods use at least 5\% of the labeled data as their most minimally supervised setting, which we argue, is an impractical level of supervision for semi-supervised NER. 
Another class of statistical 
methods \cite{peng-etal-2019-distantly, zhou-etal-2022-distantly} try to solve the reliance on gold labels by resorting to distant supervision: they construct lexicons based on large dataset independent knowledge bases. These methods then use Positive-Unlabeled (PU) learning to train classifiers using only labeled positive examples and a set of unlabeled data containing both positives and negatives.

Other approaches like \cite{liu-etal-2019-towards} showed the benefits of augmenting neural NER taggers with external gazetteers. However, external gazetteers may not always be available for particular domains. In contrast we propose a semi-supervised method for NER along the lines of the approach taken by \cite{10.5555/1621829.1621837}: one that combines the generalizability of contemporary deep learning with intuitive reasoning and linguistic insights, and we demonstrate its strong performance, under a setting of extremely low supervision. We posit that such an approach can rival more multifaceted statistical techniques for semi-supervised NER, such as PU learning \& data augmentation, among others.

\begin{figure*}[t]
    \centering
    \includegraphics[width=\textwidth]{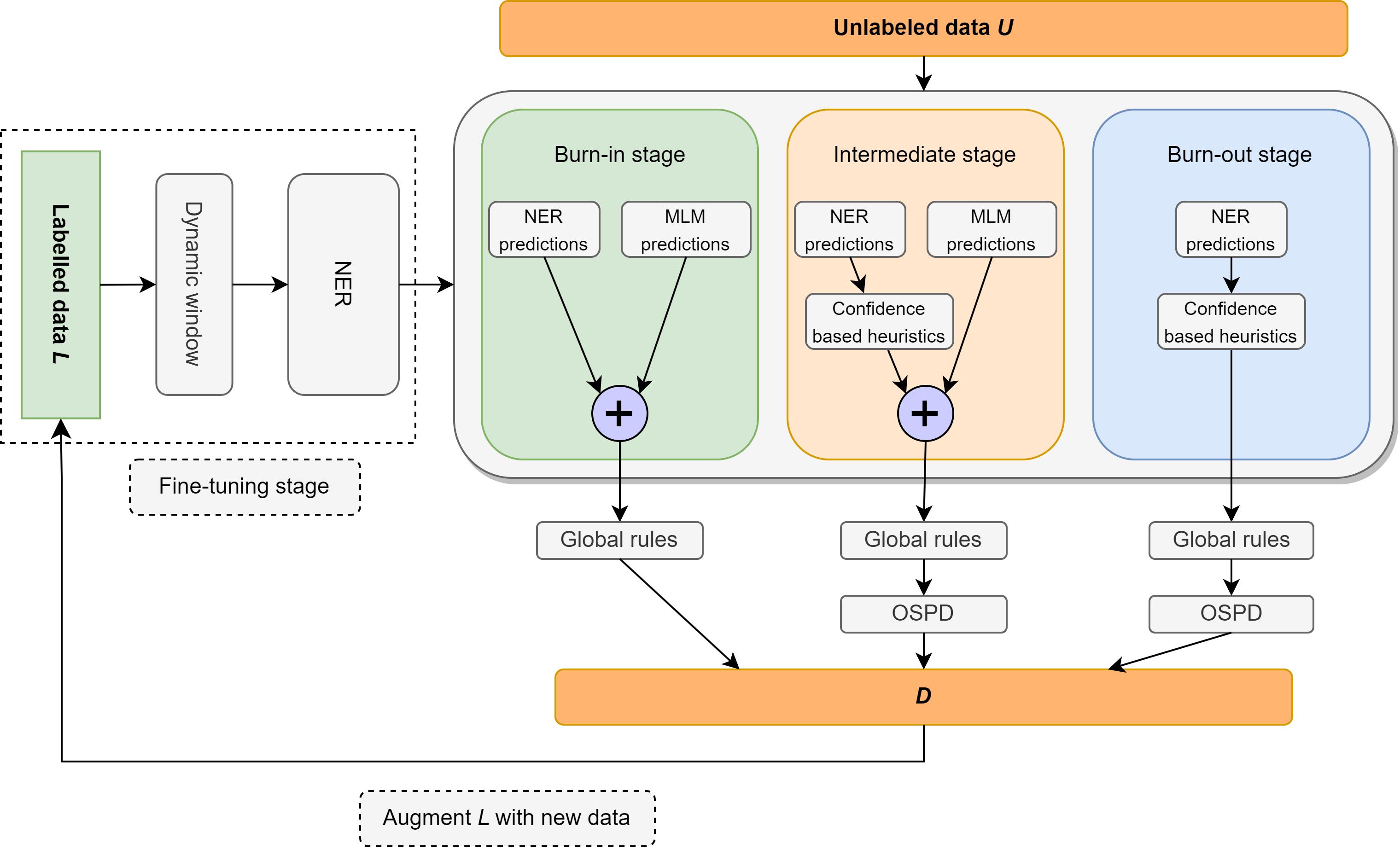} 
    \caption{\footnotesize{The proposed method illustrated. $D$ refers to a subset of the unlabeled data which is added back to the labeled data for retraining in the next iteration. 
    OSPD refers to the ``One Sense Per Discourse'' rule; ``Global rules'' indicate the rules described in Section \ref{subsec:global_rules}. The colors used in the figure represent the decreasing quality of the generated annotations in the three stages, after the fine-tuning stage: \textcolor{green}{green} $\rightarrow$ \textcolor{orange}{orange} $\rightarrow$ \textcolor{blue}{blue}.}}
    \label{fig:ellen-architecture} 
\end{figure*}

\begin{table}[h]
\centering
\small 
\renewcommand{\arraystretch}{1.5} 
\begin{tabularx}{\columnwidth}{lX}
\toprule
Category & Entities \\
\midrule
ORG & Reuters, PUK, NATO, Honda, Ajax Amsterdam, Motorola, PSV Eindhoven, PKK, Hansa Rostock, Commonwealth \\
\midrule
LOC & Germany, Australia, Britain, Spain, Italy, LONDON, Russia, China, Japan, NEW YORK \\
\midrule
MISC & Dutch, British, French, Russian, German, Iraqi, Israeli, English, Australian, American \\
\midrule
PER & Clinton, Yeltsin, Arafat, Lebed, Wasim Akram, Waqar Younis, Mushtaq Ahmed, Netanyahu, Williams, Rubin \\
\bottomrule
\end{tabularx}
\caption{\footnotesize{Lexicon generated by the domain expert following the criteria that we outline in Section \ref{subsec:subjectivity}. This lexicon serves as the seed set of entities for our model and the sole source of ``gold'' supervision.}}
\label{tab:lexicon}
\end{table}

\section{Proposed Method}\label{sec:method}
As our method does not rely on any explicitly labeled texts, we begin by discarding all labels in the NER dataset (CoNLL-2003 in this paper). We then ask a domain expert to generate a small lexicon of 10 example named entities per class, for each of the four classes in CoNLL-2003, i.e., \texttt{PER}, \texttt{ORG}, \texttt{LOC}, and \texttt{MISC}. This lexicon:
\begin{enumerate*}[label=\alph*)]
    \item is sourced entirely from the tokens in the dataset;
    \item is constructed \textit{without} looking at any of the labels in the dataset;
    \item does not rely on any external knowledge-base or dictionary; and
    \item serves as the sole source of ``gold supervision'' for our method.
\end{enumerate*}
The domain expert is able to construct the lexicon (refer to Table~\ref{tab:lexicon}) in less than 30 minutes for CoNLL-2003. We believe 10 examples per class is a reasonable number for producing an informative lexicon with the minimum amount of effort required; this is similar to few-shot relation extraction, which focuses on 5 examples/5-shots \cite{han-etal-2018-fewrel}. We then use this lexicon to annotate a small portion of the entirely unlabeled data of CoNLL-2003. In terms of ``degree of supervision'' (refer to table \ref{degree-of-supervision}), we annotate about 9.13\% of entities present in the CoNLL training data using the lexicon. We refer to this annotated subset as $L$ and the remaining unlabeled data as $U$, following the convention of traditional semi-supervised learning literature. At a high level, our approach extends the simple self-training algorithm presented by \cite{10.5555/1621829.1621837}, which is shown in Algorithm~\ref{alg:self-training-NER}. 
However, we argue that the procedure delineated in this algorithm is abstract, offering no guidance on the precise sequence in which the NER classifier $C_{k}$ (and any other linguistic rules) should be applied to extract new data $D$. Such ambiguity could potentially lead to the common pitfalls inherent in classic self-training approaches. In our study, determining the optimal order and manner in which these various approaches should be combined proved to be a nuanced task. Our work is motivated by curriculum learning, which argues that better models are learned when the training data is ``presented in a meaningful order which illustrates gradually more complex examples''~\cite{bengio2009curriculum,sachan2016easy}.
In this paper, we adjust this principle to mean {\em gradually noisier examples} based on the observation that in self-training it is critical that the initial models be of higher quality to reduce noise in future iterations.
Inspired by this idea, we propose an intuitive, three-stage framework, illustrated in Figure~\ref{fig:ellen-architecture}, for effectively combining linguistic rules with pre-trained language models such as \cite{he2023debertav3}. We design our framework to simultaneously balance two orthogonal goals:
\begin{enumerate}
\item avoiding the pitfalls of classic self-training (e.g., a model failing to correct its errors and instead amplifying them) to the extent possible, and
\item still being conceptually similar to self-training at a high level.
\end{enumerate}
Our proposed NER method is fully modular, uses the \texttt{deberta-v3-large} encoder\footnote{DeBERTa-v3 is the current state-of-the-art encoder-based model on many benchmark NLP tasks. The key advantage of using DeBERTa is its relative positional encoding, which allows the model to generalize better to longer sequences.} as the neural component, and blends various other linguistic and statistical heuristics in a sieve \cite{lee-etal-2013-deterministic}. We first describe each of these heuristics below, and then describe how they are integrated in the three-stage ELLEN framework. 

\begin{algorithm}[t]
\small
\caption{\footnotesize{A simple NER self-training algorithm}}
\label{alg:self-training-NER}
\begin{algorithmic}[1]
\State \textbf{Given:}
\State \quad \( L \) - a small set of labeled training data
\State \quad \( U \) - unlabeled data
\For{\( k \) iterations}
    \State \textbf{Step 1:} Train a NER \( C_k \) based on \( L \)
    \State \textbf{Step 2:} Extract new data \( D \) based on \( C_k \)
    \State \textbf{Step 3:} Add \( D \) to \( L \)
\EndFor
\end{algorithmic}
\end{algorithm}

\subsection{Unsupervised Entity Recognition Using A Masked Language Model (MLM)}\label{subsec:MLM}
Motivated by the observation that any language model, subjected to pretraining via the Masked Language Modeling (MLM) objective, likely acquires  semantic, syntactic, and world knowledge, we hypothesize that the capability to discern named entities is also inherently embedded within such models.
We present a novel, fully unsupervised algorithm, implemented as a rule in our neuro-symbolic modular architecture that allows us to gain additional ``free'' supervision , beyond our small lexicon of ten entities. This algorithm relies on a small pre-trained LM from \cite{DBLP:journals/corr/abs-1907-11692}, which leverages our lexicon to extract new Named Entities from unlabeled data.

Unlike recent prompt-based approaches for NER, particularly few-shot NER, which involve either prompting LLM's with in-context examples to inject NER ability \cite{chen2023learning}, or involve constructing dynamic templates based on label aware pivot words, our approach is much simpler and more constrained. We first use a very simple, linguistically inspired regular expression, based on part-of-speech (POS) tags, for detecting named entity spans: 
\begin{align*}
\texttt{(NNP|NNPS)+(IN(NNP|NNPS)+)?}
\end{align*}
where {\tt NNP}/{\tt NNPS} are the POS tags of singular/plural proper nouns, and {\tt IN} is the POS tag assigned to prepositions. On the unlabeled portion of the CoNLL training dataset, this rule can detect named entity boundaries with a precision of \textbf{85.16\%}, as shown in Table~\ref{table:regex-performance}.
Note that this rule is considerably simpler and more efficient than other recent computationally-intensive approaches, e.g., the entity typing and span identification method of \citet{shen2023promptner}, or the span classification approach adopted by \citet{arora-park-2023-split} and \citet{10.1145/3502223.3502228}. Typically, these approaches initially train a model for span classification, followed by a model for entity type classification.

\begin{table}[h]
    \centering
    \small
        \renewcommand{\arraystretch}{1.5} 
    \begin{tabular}{ccc}
        \toprule
        Precision & Recall & F1 Score \\
        \midrule
        85.16\% & 90.96\% & 87.96\% \\
        \bottomrule
    \end{tabular}
    \caption{\footnotesize{Micro-F1 scores of the regex rule for detecting named entity boundaries.}}
    \label{table:regex-performance}
\end{table}

Once entity spans are identified, we label them using a masking heuristic. Intuitively, our method selects the entity label whose exemplars in the lexicon fill in the span with the highest likelihood. More formally, we first mask the span identified by the regex above with a number of {\tt [MASK]} tokens equal to the tokens included in the span. For example, the span {\em John Doe} in the sentence: {\em John Doe is happy} will be masked as {\em {\tt [MASK]} {\tt [MASK]} is happy}. We then iteratively fill in lexicon entries of the same length (across all entity classes) and keep track of all token probabilities. For example, the entry {\em Dole} in the Person lexicon, which is tokenized as {\em Do} and {\em \#\#le},\footnote{We use the BERT tokenizer convention for multi-token words here for readability.} produces the sentence: {\em Do \#\#le is happy}. Lastly, we select the entity label based on the following formula:
$$c = \arg\max_k \max_j \frac{1}{n}\sum_i^n p(t_i| x)$$
where $t_i$ is the $i$th masked token (e.g., {\em Do} is $t_0$ in the example above); $x$ is the sentence with the masked tokens; $n$ is the total number of masks for the current example (e.g., 2 for the example above); the $j$ index iterates over all exemplars for the current entity label; and $k$ iterates over all entity class lexicons. That is, for each exemplar, we first compute the average probability of all its tokens. Then we pick the exemplar with the highest probability in a given lexicon as the probability of the corresponding entity label. Lastly, we select the entity label $c$ with the highest probability. 

We denote this rule as the Masked Language Modeling (MLM) Heuristic. We demonstrate its NER effectiveness on the development set of CoNLL-2003, in a fully unsupervised fashion in Table~\ref{table:mlm-conll-dev-performance}. As shown in the table, the F1 score of the MLM heuristic is over 56\% on the development set of CoNLL-2003. In each iteration of the procedure shown in figure \ref{alg:self-training-NER}, the MLM is used in the burn-in and intermediate stages to annotate a subset of the unlabeled data $U$, which is eventually added back to $L$.

\begin{table}[ht]   
    \centering
    \renewcommand{\arraystretch}{1.5} 
    \small
    \begin{tabular}{lccc}
        \toprule
        Entity Type & Precision & Recall & F1 \\
        \midrule
        Overall & 61.78\% & 51.90\% & 56.41\% \\
        LOC & 69.72\% & 41.53\% & 52.05\% \\
        MISC & 45.18\% & 55.15\% & 49.67\% \\
        ORG & 44.85\% & 40.88\% & 42.77\% \\
        PER & 85.07\% & 65.02\% & 73.71\% \\
        \bottomrule
    \end{tabular}
    \caption{\footnotesize{P/R/F1 of the Masked Language Modeling Heuristic as a fully unsupervised NER algorithm on the CoNLL-2003 development set. It obtains an entity-level F1 score of 56.41\% with over 60\% overall precision.}}
    \label{table:mlm-conll-dev-performance}
\end{table}

\subsection{Dynamic Window Filtering}
In most lightly supervised settings, NER models tend to suffer from the ``unlabeled entity problem'' as described in \cite{li2021empirical}, where the entities of a sentence may not be fully annotated. This tends to seriously degrade model performance, since the model treats unlabeled entities as negative or \texttt{O}/Outside instances. Even self-training methods may not be sufficient to completely alleviate the false negative problem since they are susceptible to confirmation bias \cite{9207304}, i.e., erroneously predicted pseudo-labels are likely to deteriorate the model’s performance in subsequent rounds of training. In contrast to \cite{li2021empirical}'s method which uses negative sampling to avoid training the NER on unlabeled entities, we propose a very simple and run-time efficient linguistically inspired algorithm for controlling the effect of false negatives in sparsely annotated data settings, such as ours. We refer to our algorithm as ``Dynamic Window Filtering.'' Using part-of-speech (POS) tag information\footnote{POS tags are now available for many languages (see \url{https://universaldependencies.org/}) and can be obtained from various off-the-shelf models or language processing tools.} and an intuition that tokens which are labeled as \texttt{O}/Outside and which possess a POS tag of {\tt NNP} (singular proper noun) are highly likely to be unlabeled named entities and thus should be discarded from the NER's training data. We implement this algorithm as the following rule: we slide a contextual window across each sentence in the labeled subset of the data, $L$, and for each named entity segment we encounter whose label is known, we create a window of size $W$, which dynamically expands in both directions around the labeled entity until an \texttt{O} token that is also tagged with the POS tag of {\tt NNP} (singular proper noun), is encountered\footnote{We note that alternatives to Dynamic Windowing exist for managing unlabeled entities. For instance, as demonstrated in \cite{vacareanu-etal-2024-active}, unlabeled entities can remain in the NER's training data with their impact mitigated by excluding them from the loss calculation, i.e., by backpropagating only over gold-annotated tokens.}. We also emphasize that our POS tags are inherently noisy, since they are obtained from an external LSTM-CRF based POS tagger that was trained exclusively on the Penn Treebank corpus \cite{marcus-etal-1993-building}. We \textit{do not} use the gold POS tags from the CoNLL data. The example below illustrates this rule:\\\\
\noindent \textbf{Example 1:} \\
\textit{\textbf{EU} rejects \textbf{German} call to boycott \textbf{British} Lamb.}\\

\noindent Suppose ``EU'' and ``British'' are both Named Entities with known labels and are also proper nouns. Suppose ``German'' is also a proper noun but with an unknown named entity label (and, thus, it is currently labeled as {\tt O}).  Dynamic Window Filtering creates a contextual window around ``EU'' and ``British'', expanding in size until it encounters the token ``German''. This algorithm would thus break the original example into two new segments: 
\begin{enumerate}
    \item ``\textbf{EU} rejects''
    \item ``call to boycott \textbf{British} lamb.''
\end{enumerate}

In every stage of our method, we apply dynamic window filtering on the data that the NER is trained on. This includes both the initial set of sparsely annotated gold data and its augmentations with the pseudo-labeled data ($D$) that is extracted from the unlabeled data ($U$) in each iteration. We find that this algorithm achieves the same goal as the method presented in \cite{li2021empirical} i.e. discarding Named Entities with unknown labels, while being much simpler and computationally cheaper.

\subsection{Global Rules}\label{subsec:global_rules}
Lightly supervised NER models may confuse named entities between Organizations and Persons, Organizations and Locations (and vice versa) due to their shared context. To remedy this, we apply a series of commonsense linguistic rules on the aggregated predictions of the NER model and the Masked Language Modeling (MLM) heuristic. We apply rules for disambiguating named entity segments that have been tagged as Persons (\texttt{PER}) but end in a company suffix. We update the labels of such segments, including the company suffix to \texttt{ORG}. For example, if the entity segment ``Walt Disney'' is tagged as \texttt{PER}, but it is immediately followed by ``Inc.'' (a company suffix), the rule would force the whole segment ``Walt Disney Inc.'' to be an \texttt{ORG}. Similarly, if any named entity segment is tagged as a Location, but it is followed or follows a segment tagged as an Organization, we update the labels of the both the Location segment and the Organization segment to be \texttt{ORG}. 

Additionally, we observe that many instances of the CoNLL-2003 validation set consist of terse reports of scores of games between sports teams (which are Organization entities), but which also semantically overlap with Location names. For example, the name ``Somerset'' could refer to a county in England (\texttt{LOC}) or a cricket club (\texttt{ORG}). It is common for a lightly supervised model to confuse the labels to be assigned to such examples. To remedy this, we propose an additional heuristic, which identifies segments labeled as Locations (\texttt{LOC}) and if these segments are followed a score token or at least two integer numbers resembling a score\footnote{We use regular expressions to detect score tokens, integer patterns, and hyphens respectively.}, we force their labels to be \texttt{ORG}. 

\subsection{One Sense Per Discourse}
\cite{amalvy-etal-2023-role} demonstrated the significance of both local and global document-level context in enhancing the efficacy of pre-trained transformer-based models for NER. In our work, we harness the document-level metadata provided in CoNLL-03 to integrate the ``One Sense Per Discourse'' (OSPD) principle \cite{gale1992one} into our neuro-symbolic approach. Primarily conceived for word sense disambiguation, OSPD posits that a term's sense remains consistent when repeatedly used within a cohesive discourse. We operationalize this idea by asserting that if a named entity's predominant classification within a CoNLL discourse leans towards a particular label, then all instances of that entity within the discourse should adopt this dominant label. For instance, should ``IBM'' appear five times in a document—thrice as an \texttt{ORG} and twice as a \texttt{LOC}—our method dictates that all mentions of ``IBM'' be labeled as \texttt{ORG} due to its majority occurrence.

\subsection{Confidence-Based Rules}
In semi-supervised learning, classifier confidence can effectively guide the inclusion of unlabeled data. Nonetheless, contemporary deep neural networks often produce overconfident predictions. To harness the confidence-based heuristics outlined in \cite{10.5555/1621829.1621837}, we adopt the ``Smoothed Generalized Cross Entropy'' loss from \cite{zhang2018generalized} \& \cite{Dimachkie_2023}, which has been shown to regulate and calibrate model predictions. We then include the following rules in our method: For any segment of tokens classified as an \texttt{ORG}, \texttt{LOC}, or \texttt{PER} with a classifier confidence score \( > T \)\footnote{In our experiments, we empirically set \( T \) to 0.9.}, we find other mentions of the same segment within the same CoNLL document and force their label(s) to be the same as the high confidence segment. This is known as the \textit{multi-mention} heuristic. In addition, if the high confidence segment ends in a company suffix, we remove the company suffix and apply the multi-mention property on the remaining segment. We apply the same rule for a high confidence segment that begins with a Person title (from a list of common English honorifics). Furthermore, for each segment ending in a company suffix or starting with a Person prefix, we remove the affix, while retaining the context, to form a new, previously unseen sentence which we then reclassify. For example, suppose we have a sentence in the training data with a \texttt{PER} segment tagged with high confidence as follows: ``\textit{The meeting was led by \textbf{Ms. Taylor}}.'' Then, removing the Person title would yield the new sentence: ``\textit{The meeting was led by \textbf{Taylor}}.'' Should the predicted labels of this altered segment in the new sentence differ from those before the affix removal, especially if classified without high confidence, we designate such sentences for inclusion in subset $D$. This subset is reintroduced to the training data in the subsequent semi-supervised learning iteration, as illustrated in figure \ref{fig:ellen-architecture}. We apply these confidence-based heuristics in a sieve i.e. in order of decreasing precision.
\vspace{-2mm}
\subsection{Minimizing The Dependency On A Lexicon}\label{subsec:subjectivity}
We minimize the dependency of our self-training algorithm on the lexicon chosen by the domain expert by outlining a process that they must follow for picking the lexicon. Using the simple regular expression defined in Section \ref{subsec:MLM} (that is able to detect named entity boundaries with high precision), we harvest named entity candidates from the CoNLL-2003 training data in a fully unsupervised manner. These candidates, ranked by their frequency of occurrence in the data, are presented to the domain expert who is tasked to select the most frequent and \textit{unambiguous} ones for each class (i.e. for each class, the lexicon should not contain entities that overlap with another class). By enforcing this criteria of objectivity, we implicitly minimize the chances of multiple domain experts picking vastly different lexicons, thereby minimizing the effects of lexicon variability on our method.
\subsection{ELLEN: Integrating Neural And Symbolic Components}
In Figure~\ref{fig:ellen-architecture}, we depict a three-stage framework for amalgamating the heuristics and determining the subset $D$ from unlabeled data $U$ to augment the labeled set $L$ in each semi-supervised learning iteration. This procedure involves three stages: initial burn-in, subsequent intermediate stage, and a concluding burn-out stage. Our selection criterion for $D$ is straightforward: only sentences with entity label updates due to heuristics are considered. This approach aims to expand model knowledge within each cycle while curtailing classic self-training pitfalls. During burn-in, predictions from the Masked Language Modeling Heuristic (MLM) are combined with those from the NER, favoring the MLM due to the NER's initial weakness. Confidence-based heuristics, reliant on the NER's outputs, are deferred. Global rules and OSPD are applied solely on sentences modified by the MLM. As the NER matures through training on MLM outputs, the intermediate phase gives it precedence over the MLM; here, confidence-based heuristics solely target NER predictions, while global rules and OSPD extend to any NER or model-updated sentence. In the burn-out phase, we relax constraints, allowing full self-training. With the model now robust, it gleans any residual data from $U$ for a final training iteration.

\section{Experimental Results}
\subsection{Data \& Setup}
We evaluate our method on CoNLL-2003 using three different degrees of supervision (see Table \ref{degree-of-supervision}). We define the ``\textbf{degree of supervision}'' to be the percentage of named entities annotated, relative to the total number of entities present in the data. The first setting is the proposed extremely lightly supervised setting, equivalent to 9.13\% in terms of degree of supervision or about ``1\%'' in terms of the number of labeled sentences. We borrow the second ``5\%'' data setting (which corresponds to the first 700 sentences in CoNLL-03's training split) from the current state-of-the-art approaches on semi-supervised NER \cite{chen-etal-2020-local,zheng-etal-2023-jointprop}.  However, unlike \citet{chen-etal-2020-local}, which uses \texttt{Fairseq} \cite{ott2019fairseq} for augmenting the unlabeled data with equivalent back-translations from German, we sample 10,000 unlabeled sentences at random\footnote{We do 3 random augmentations for choosing the 10,000 unlabeled sentences. The first 700 sentences are chosen without randomization, to keep the data setting exactly the same as Semi-LADA \cite{chen-etal-2020-local}  \& JointProp \cite{zheng-etal-2023-jointprop}.}, without any augmentation. The third setting is the fully supervised setting, where we evaluate the effectiveness of ELLEN against ACE \cite{wang2020automated}, the current SOTA method on CoNLL-03, and a DeBERTa V3 \cite{he2023debertav3} classifier\footnote{\url{https://huggingface.co/tner/deberta-v3-large-conll2003}} finetuned on CoNLL-03.

To summarize, the three different sources of supervision in our experiments, are as follows:
\begin{enumerate}
    \item \textbf{1\% data setting:} We use the unambiguous lexicon produced by the domain expert consisting of 10 examples for each of the four CoNLL-03 classes: \texttt{MISC}, \texttt{ORG}, \texttt{LOC} and \texttt{PER}.
    \item \textbf{5\% data setting:} In this setting, we also extract an unambiguous lexicon from the entities within the first 700 sentences of the CoNLL-03 training split, adhering to the consistent definition of `unambiguous' as described in Section \ref{subsec:subjectivity}—entities within each class must not overlap with those from other classes. This lexicon, comprising 98 examples for the \texttt{MISC} class, 174 for \texttt{ORG}, 189 for \texttt{LOC}, and 274 for \texttt{PER}, is then used for annotating the unlabeled data and for the MLM.
    \item \textbf{Fully supervised setting:} In this setting, we \textit{do not} use a lexicon for annotating the data since all gold labels are available. However, since the MLM heuristic (section \ref{subsec:MLM}) requires a lexicon, we extract an unambiguous one just for the MLM, from all of the labeled sentences in CoNLL-03's training split. This lexicon contains 868 examples for the \texttt{MISC} class, 2329 for \texttt{ORG}, 1245 for \texttt{LOC}, and 3598 for \texttt{PER} (refer to Appendix \ref{sec:appendix3}).
\end{enumerate}

\begin{table}[h]
\centering
\small
\renewcommand{\arraystretch}{1.5} 
\footnotesize
\begin{tabular}{lccc}
\toprule
Setting  & 1\% & 5\% & Supervised\\
\midrule
Supervision degree & 9.13\% & 28.5\% & 100\% \\
\midrule
\# Labeled tokens & 2569 & 6971 & 34043 \\
\bottomrule
\end{tabular}
\caption{\footnotesize{Statistics on degrees of supervision used in this work. 5\% (in terms of number of sentences) is a common setting for semi-supervised NER. For the 1\% and 5\% settings, we calculate the supervision degree based on unambiguous lexicons.}}
\label{degree-of-supervision}
\end{table}

\subsection{Results Using 1\% Labeled Data}
As shown in table \ref{extreme-performance}, in the extremely lightly supervised setting, which is more restrictive than typical semi-supervised NER approaches, we find that our method achieves an F1 score of \textbf{76.87\%} on the CoNLL-2003 test set. The only supervision here comes from a domain expert's lexicon which itself does not use any gold labels from CoNLL-2003. This result indicates our method's real-world effectiveness, where annotations are scarce and lexicons like ours are the only source(s) of supervision.

\begin{table}[h]
\centering
\small
\setlength{\arrayrulewidth}{0.5mm}
\renewcommand{\arraystretch}{1.5}
    \begin{tabularx}{\columnwidth}{p{0.85cm}X X X} 
        \toprule
        Type & Precision & Recall & F1 \\
        \midrule
        Overall & 74.63 \scriptsize{$\pm 0.33\%$} & 79.26 \scriptsize{$\pm 0.92\%$} & 76.87 \scriptsize{$\pm 0.48\%$} \\
        LOC & 87.92 \scriptsize{$\pm 1.21\%$} & 78.68 \scriptsize{$\pm 4.76\%$} & 83.04 \scriptsize{$\pm 2.36\%$} \\
        MISC & 56.32 \scriptsize{$\pm 1.82\%$} & 61.00 \scriptsize{$\pm 1.25\%$} & 58.57 \scriptsize{$\pm 0.53\%$} \\
        ORG & 62.29 \scriptsize{$\pm 0.39\%$} & 77.26 \scriptsize{$\pm 1.64\%$} & 68.97 \scriptsize{$\pm 0.51\%$} \\
        PER & 87.98 \scriptsize{$\pm 0.92\%$} & 92.71 \scriptsize{$\pm 0.27\%$} & 90.28 \scriptsize{$\pm 0.43\%$} \\
        \bottomrule
    \end{tabularx}
\caption{\footnotesize{Precision/Recall/F1 scores for ELLEN on CoNLL-2003 under the extremely lightly supervised setting. All of our runs are averaged over 3 random seeds. We present entity-level metrics using the official CoNLL-scoring script.}}
\label{extreme-performance}
\end{table}

\subsection{Results Using 5\% Labeled Data}
Under the 5\% data setting (or 28.5\% in terms of ``degree of supervision''), we show that our method achieves an F1 score of \textbf{84.87}\% (Table \ref{five-percent-performance}), outperforming more complex methods like PU learning \cite{zhou-etal-2022-distantly}, models based on hierarchical latent variables \cite{chen2019variational} \& those employing noise strategies \cite{lakshmi-narayan-etal-2019-exploration}. We find that it also performs favorably compared to Semi-LADA \cite{chen-etal-2020-local} without using any back-translations for the unlabeled data. More importantly, we highlight that our method outperforms PU-Learning approaches whilst using much fewer exemplars per class (the PU-Learning methods of \cite{peng-etal-2019-distantly} \& \cite{zhou-etal-2022-distantly} rely on a lexicon that contains ``2,000 person names, 748 location names, 353 organization names, and 104 MISC entities''). We include Table \ref{tab:PU_learning_lexicon} (Appendix \ref{sec:appendix1}), which is directly taken from \cite{peng-etal-2019-distantly}, to illustrate this. 

Furthermore, we also highlight Figure 4 from GPT-NER \cite{wang2023gptner} which shows the performance of ACE \cite{wang2020automated} in a low-resource context. Specifically, when ACE is trained on a 1\% subset (in terms of the number of sentences) of CoNLL-03's training data, it's F1 score falls below 20\% and below 70\% when trained on a 5\% subset. Although a fair comparison with our method cannot be made due to differing definitions of ``low resource,'' it is noteworthy that ELLEN attains F1 scores of 76.87\% and 84.87\% under our equivalent ``1\%'' and ``5\%'' settings respectively, suggesting that our method significantly outperforms ACE in resource-constrained scenarios.

\begin{table}[h] 
    \centering
    \small
    
\renewcommand{\arraystretch}{1.5}
    \begin{tabularx}{\columnwidth}{Xp{1.3cm}p{1.3cm}p{1.3cm}} 
        \toprule
        Methods & P & R & F1 \\
        \midrule
        VSL-GG-Hier              & 84.13\% & 82.64\% & 83.38\% \\
        MT + Noise               & 83.74\% & 81.49\% & 82.60\% \\
        Semi-LADA                & 86.93\% & 85.74\% & 86.33\% \\
        Jointprop       & \textbf{89.88\%} & \textbf{85.98\%} & \textbf{87.68\%} \\
        PU-Learning              & 85.79\% & 81.03\% & 83.34\% \\
        ELLEN†                   & 81.88 \scriptsize{ $\pm 1.18\%$ }\ & 88.01 \scriptsize{$\pm 0.19\%$} & 84.87 \scriptsize{$\pm 0.62\%$} \\
        \bottomrule
    \end{tabularx}
    \caption{\footnotesize{Performance on CoNLL 2003 with 5\% labeled data. It should be noted that JointProp \cite{zheng-etal-2023-jointprop} is a multi-task learning framework. All of our runs are averaged over 3 random seeds.}}
    \label{five-percent-performance}
\end{table}

\subsection{Zero-Shot Evaluation}
We apply our extremely lightly supervised (``1\%'') method in a zero-shot manner on WNUT-17, a dataset from the social media domain, characterized by noisy text. 
That is, using the model that was trained on CoNLL-03 with only a lexicon of 10 samples per class (provided by the domain expert), we proceed to evaluate this model on the WNUT-17 test dataset.
After aligning the predictions of each model with the label space of CoNLL-03 (see Appendix \ref{sec:appendix2} for details), we observe that ELLEN achieves comparable zero-shot performance to GPT-3.5 and GPT-4, and also achieves relatively strong zero-shot performance against a {\em fully supervised} model\footnote{We use the RoBERTa large model, available here: \url{https://huggingface.co/tner/roberta-large-wnut2017}} from the the T-NER library \cite{ushio-camacho-collados-2021-ner} that was actually trained on WNUT-17 gold data (see Table \ref{tab:zero-shot}). This result is exciting because it indicates the potential for our method to be used across domains, given the relatively small size of our model and its extremely light supervision.

\begin{table}[h]
\centering
\small
\renewcommand{\arraystretch}{1.5} 
\begin{tabularx}{\columnwidth}{@{}l *{6}{>{\RaggedRight\arraybackslash}X}@{}}
\toprule
Method & \multicolumn{1}{l}{LOC} & \multicolumn{1}{l}{MISC} & \multicolumn{1}{l}{ORG} & \multicolumn{1}{l}{PER} & \multicolumn{1}{l}{AVG} \\
\midrule
T-NER & \textbf{64.21\%} & \textbf{42.04\%} & \textbf{42.98\%} & \textbf{66.11\%} & \textbf{55.11\%} \\
GPT-3.5 & 49.17\% & 8.06\% & 29.71\% & 59.84\% & 39.96\% \\
GPT-4 & 58.70\% & 25.40\% & 38.05\% & 56.87\% & 43.72\% \\
ELLEN† & 44.82 \scriptsize{$\pm 3.84\%$} & 6.21 \scriptsize{$\pm 1.25\%$} & 26.49 \scriptsize{$\pm 5.01\%$}  & 67.00 \scriptsize{$\pm 3.54\%$} & 41.56 \scriptsize{$\pm 0.92\%$} \\
\bottomrule
\end{tabularx}
\caption{\footnotesize{Comparing F1 scores: ELLEN, GPT-3.5, and GPT-4 are evaluated in zero-shot mode against T-NER's fully supervised model on the WNUT-17 test set, after label alignment with CoNLL-03 († indicates our framework). For ELLEN, we report the average zero-shot score of 3 different random initializations and training runs of the models under extremely light supervision.}}
\label{tab:zero-shot}
\end{table}

\subsection{Results Using Full Supervision}
Lastly, we show that our method can also be adapted to a fully-supervised setting. Table \ref{tab:fully-supervised} shows that we obtain a respectable F1 score of \textbf{90.98\%} relative to ACE (\textbf{94.6\%}) and a standard supervised classifier (\textbf{92.2\%}). We note that, while our neuro-symbolic approach is effective in low resource settings, when full supervision is available, other methods may outperform ours. This is primarily due to the noise introduced by the various heuristics we propose in Section \ref{sec:method} (MLM, One Sense Per Discourse, confidence-based rules), which may erroneously annotate \texttt{O}/Outside entities as belonging to a non-\texttt{O} class, leading to our model being iteratively retrained on some noisy data (as shown in Figure \ref{fig:ellen-architecture}).

\begin{table}[h]
\centering
\small
\renewcommand{\arraystretch}{1.5} 
\begin{tabular}{lccc}
\toprule
Model       & F1 \\
\midrule
ELLEN       & 90.98 \scriptsize{$\pm 0.54\%$}\\
DeBERTa V3  & 92.2\%\\
ACE \cite{wang2020automated} & \textbf{94.6\%}\\
\bottomrule
\end{tabular}
\caption{\footnotesize{Performance of ELLEN on CoNLL-2003 test when using all available annotations from the CoNLL-2003 training data (fully supervised setting). For ELLEN, we report an average of training runs over 3 random seeds.}}
\label{tab:fully-supervised}
\end{table}

\subsection{Error Analysis And Ablation Experiment}
Focusing on the extremely lightly supervised setting for CoNLL-03, we observed that over 30\% of model errors on validation data involve confusing \texttt{ORG} and \texttt{LOC} entities. These errors can be attributed to a combination of factors: a) a bias in CoNLL-03's validation and test data towards sporting events not sufficiently reflected in the training data, b) the inadequacy of a global rule (refer to Section \ref{subsec:global_rules}) to differentiate between sports teams (\texttt{ORG}) and locations (\texttt{LOC}) in nuanced contexts, e.g., both `YORKSHIRE' and `HEADINGLEY' would be labeled as \texttt{ORG}'s in the sentence: ``YORKSHIRE AT HEADINGLEY'' even though `HEADINGLEY' is a \texttt{LOC}. c) errors arising from noisy POS tags and incorrectly identified entity spans, e.g., ``Dhaka Stock Exchange'' would be identified as two separate entities ``Dhaka'' and ``Stock Exchange'' by the regular expression, leading to incorrect labels by the MLM Heuristic during training; and d) confusion between \texttt{ORG} and \texttt{MISC} classes, partly because the \texttt{MISC} class lexicon primarily includes nationalities, which does not fully represent its broader scope (e.g., events, products, works of art).

In an ablation on the CoNLL-2003 dev set (Table \ref{tab:ablation-study}), we found the MLM to be the most impactful component. This was followed by dynamic window filtering, which allows our method to achieve a \textbf{64.7\%} F1 score on its own. Importantly, reintegrating other components—OSPD, confidence-based heuristics, and global rules—each further enhances performance, underscoring their collective contribution to the method's effectiveness.

\begin{table}[h]
\centering
\renewcommand{\arraystretch}{1.5} 
\setlength{\tabcolsep}{4pt} 
\label{ablation-studies}
\small
\begin{tabular}{lccc}
\toprule
Ablations & P & R & F1 \\
\midrule
Full system & 71.31 \scriptsize{$\pm 2.4\%$}& 75.90 \scriptsize{$\pm 0.9\%$} & 73.52 \scriptsize{$\pm 1.6\%$} \\
MLM & 61.96 \scriptsize{$\pm 1.1\%$}& 74.00 \scriptsize{$\pm 1.0\%$} & 67.40 \scriptsize{$\pm 0.2\%$} \\
CR, GR, & 70.30 \scriptsize{$\pm 3.5\%$} & 74.36 \scriptsize{$\pm 0.8\%$} & 72.23 \scriptsize{$\pm 2.1\%$} \vspace{-2mm}\\OSPD \\
MLM, CR, & 59.20 \scriptsize{$\pm 1.8\%$} & 71.30 \scriptsize{$\pm 1.0\%$} & 64.70 \scriptsize{$\pm 1.3\%$} \vspace{-2mm}\\GR, OSPD \\
\bottomrule
\end{tabular}
\caption{\footnotesize{Ablation of major components in our system, measured by P/R/F1 on CoNLL-03 validation data. \textbf{`CR'} refers to ``Confidence-Based Rules.'' \textbf{`GR'} refers to ``Global Rules.'' \textbf{`OSPD'} refers to ``One Sense Per Discourse''. \textbf{`MLM'} refers to the ``Masked Language Model''.}}
\label{tab:ablation-study}
\end{table}

\section{Conclusion}
In this paper, we present a framework that harmoniously blends linguistics and deep learning to overcome the paucity of labeled data for NER, requiring significantly less supervision than previous methods. Real-world entity extraction is often hindered by the lack of annotated data, especially in low-resource domains. While LLMs offer potential remedies, they are not without limitations. Our solution, ELLEN, introduces an efficient, encoder-only method that enables the assembly of an NER system in as little as ``half a day'', requiring only a single expert-provided lexicon. We show ELLEN's strong performance in the extremely low resource setting, showing that it scales well under varying supervision levels, while also outperforming other, more complex approaches.

\section*{Acknowledgements}
This work was partially supported by the
Defense Advanced Research Projects Agency
(DARPA) under the Habitus
program and by the National Science Foundation (NSF) under grant \#2006583. Mihai Surdeanu declares a
financial interest in lum.ai. This interest has
been properly disclosed to the University of
Arizona Institutional Review Committee and
is managed in accordance with its conflict of
interest policies.

\section*{Limitations}
Our proposed method, while showcasing promising results in settings of lightly supervised named entity recognition (NER), faces certain limitations that warrant discussion. Primarily, our evaluation was conducted only on two flat NER datasets. Adapting our method across a broader spectrum of datasets, especially those that may feature more complex, fine-grained, or nested entity structures, needs further exploration. Consequently, our current approach does not explicitly address the challenges associated with more intricate NER tasks, such as nested, fine-grained, hierarchical or intersectional NER, which require the identification of entities within entities or the recognition of novel entity types beyond traditional categories.

Some of the rules employed by our method are domain and language-specific, which could limit their wider applicability. However, we also highlight that four out of the eleven total rules in our method are domain and language-independent (assuming the existence of a language model for that domain/language). These include the Masked Language Model (MLM), a heuristic for ``free'' supervision from exemplars (which can come from any domain or language), Dynamic Window Filtering, which assumes that POS tags are available for a given language/domain and that the language distinguishes between common nouns and proper nouns, One Sense Per Discourse (OSPD) which simply propagates the majority label within a document, and the basic multi-mention heuristic for label propagation which only uses classifier confidence scores. MLM and dynamic window filtering, both language and domain-independent, are the two components that contribute the most to the performance of our NER method (as shown in Table \ref{tab:ablation-study}).

Certain rules, such as those dependent on company suffixes and person honorifics, are not domain-independent but are transferable across languages with the adaptation of language-specific affixes. This adaptability suggests a pathway to applying our method to new languages, provided a list of relevant suffixes and honorifics is used. However, there may be challenges in directly applying some of these rules to languages which use vastly different conventions for naming entities. Nevertheless, in presenting our findings, we have not claimed our method to be universally applicable across all languages and domains. Instead, we aimed to demonstrate how the integration of linguistic insights with neural networks can mitigate the scarcity of labeled data in NER. Our framework, which harmoniously blends these elements, points to a significant step forward, while highlighting the necessity for further research to extend its applicability to more diverse and complex NER scenarios.

\section*{Ethical Considerations}
This work utilizes two public, commonly-used datasets for Named Entity Recognition (NER). One dataset, WNUT-17, derived from the social media domain, mainly consists of user-generated comments, of which a very small portion may include language that some might find inappropriate or offensive. Furthermore, our approach incorporates open-source pre-trained language models. Thus, any biases inherent in these models due to their pre-training data would also apply to our work. Regarding the selection of named entity seeds (see Section \ref{subsec:subjectivity}), while efforts were made to minimize subjectivity in the creation of the lexicon, it is theoretically possible for the proposed method to be used to intentionally introduce biases into an NER model. However, we believe that, apart from the potential issues already mentioned, this work does not raise any significant ethical concerns.

\section{Bibliographical References}\label{sec:reference}

\bibliographystyle{lrec-coling2024-natbib}
\bibliography{lrec-coling2024-example}

\section{Language Resource References}
\label{lr:ref}
\bibliographystylelanguageresource{lrec-coling2024-natbib}
\bibliographylanguageresource{languageresource}

\appendix
\section{Statistics of Labeling With PU Learning Lexicon}
\label{sec:appendix1}
\begin{table}[h]
\setlength{\tabcolsep}{4.6pt} 
    \centering
    \small
    \renewcommand{\arraystretch}{1.5} 
    \begin{tabular}{l l r r r}
        \toprule
        Type & \# of 1.w. & Precision & Recall \\
        \midrule
        PER  & 2,507 & 89.26\% & 17.38\% \\
                   LOC  & 4,384 & 85.07\% & 50.03\% \\
                   ORG  & 3,198 & 86.17\% & 29.45\% \\
                   MISC & 1,464 & 92.13\% & 30.59\% \\
        \bottomrule
    \end{tabular}
    \caption{\footnotesize{Data labeling results using the lexicon used by PU Learning methods: the number of labeled words (\# of l.w.), the word-level precision (\# of true labeled words/\# of total labeled words) and recall, on CoNLL-2003.}}
    \label{tab:PU_learning_lexicon}
\end{table}
Table \ref{tab:PU_learning_lexicon}, which is directly taken from the work of \cite{peng-etal-2019-distantly}, illustrates the data labeling statistics of the large lexicons (sourced from external dictionaries) used by PU-learning methods. This is in stark contrast to our method, where the lexicon contains only 10 exemplars per class in the `1\%' data setting or at most, a few hundred for the \texttt{PER} class in the `5\%' data setting.

\section{WNUT-17: Zero-Shot Evaluation}
\label{sec:appendix2}
To allow our model, which was trained on CoNLL-2003, to be fairly compared in a zero-shot setting against the fully supervised \texttt{roberta-large} model from the T-NER \cite{ushio-camacho-collados-2021-ner} library (which is trained on WNUT-17 gold data), we mapped the generated labels from the fully supervised model onto the label space of CoNLL-2003. We aligned the classes from WNUT-17 with CoNLL classes (\texttt{ORG, LOC, PER, MISC}) based on semantic overlap. We aligned the ‘products’ and ‘creative-work’ classes with the MISC class from CoNLL. This is because the MISC class from CoNLL also contains many product names and `works of art,' e.g., ``Ain't No Telling'' by Jimi Hendrix.

We aligned the `group' class from WNUT-17 with `ORG' from CoNLL because many `group` names in the WNUT-17 test data have a semantic overlap with organizations, e.g., `Nirvana’', `San Diego Padres.' Based upon our inspection of the data, the `group' class also includes entities like musical bands, sports teams, non-profit organizations, political groups, etc. Such entities fit well within the typical CoNLL understanding of an `organization.' For the remaining classes of WNUT-17 (\texttt{corporation, location, person}), we mapped them directly to their corresponding CoNLL-03 equivalents. We also applied this mapping to the zero-shot predictions of GPT-3.5 and GPT-4, to allow all models to be fairly compared against each other. We evaluated all models shown in Table \ref{tab:zero-shot} on the full test set (1287 samples) of WNUT-17.

We accessed GPT-3.5 and GPT-4 through the Azure OpenAI service, using the \texttt{gpt-35-turbo-0613} and \texttt{gpt-4-0613} models with \texttt{temperature=0} for deterministic results. We borrow the prompt format from the vanilla zero-shot prompt used by \cite{xie2023empirical} (shown in the figure below). In the zero-shot evaluation with GPT-3.5 and GPT-4, we observed issues similar to those observed by \cite{wang2023gptner}, i.e., both LLMs often fail to match the output length with the input sentence's token count in sequence labeling tasks like NER, a challenge amplified in longer sentences. This is documented in Table \ref{table:error_comparison}, distinguishing ``Misalignment errors''—the discrepancy in the number of LLM generated NER tags versus sentence tokens—and ``Parsing errors,'' where the LLM generation does not form a valid sequence of NER labels and hence, cannot be parsed, with GPT-3.5 showing more pronounced issues.

To mitigate these alignment problems, we used a simple approach: 
\begin{enumerate}
    \item For outputs with fewer NER tags than input tokens, we padded the sequence on the right with `\texttt{O}' tags to equalize the lengths.
    \item For outputs with excess NER tags, we truncated the surplus from the right to match the input token sequence length.
\end{enumerate}

We then evaluated the aligned and corrected predictions of both LLM's on the WNUT-17 test set using the official CoNLL-scoring script (results reported in Table \ref{tab:zero-shot}).

\begin{table}[H]
\centering
\small
\renewcommand{\arraystretch}{1.5} 

\begin{tabular}{lcc}
\toprule
Model & Misalignment Errors & Parsing Errors \\ \midrule
GPT-3.5  & 426                      & 80                                    \\
GPT-4    & 195                      & 44                                    \\ \bottomrule
\end{tabular}
\caption{\footnotesize{Comparison of error counts between GPT-3.5 and GPT-4.}}
\label{table:error_comparison}
\end{table}





\begin{tcolorbox}[colback=white,colframe=blue!75!black,title={Prompt Used For Zero-Shot Evaluation of GPT-3.5/4}]
Given entity label set: [`B-PER', `I-PER', `B-ORG', `I-ORG', `B-LOC', `I-LOC', `B-MISC', `I-MISC', `O']\\
Based on the given entity label set, please recognize the named entities for each token in the given text, and return the answer as a list of named entity tags.\\
Text: \{input text\}\\\\
Answer: \{ChatGPT response\}
\end{tcolorbox}

\section{Masked Language Model (MLM): Inverse Breaking Ties}
\label{sec:appendix3}
In order to obtain more robust annotations from the Masked Language Model (MLM), we only consider the entity span $x_i$ labeled by the MLM where the difference between the score of the class that is predicted with the highest probability and the score of the class that is predicted with the second highest probability is greater than some threshold $t_{class}$:

$$x_i \mid P(y_i = l_1 | x_i) - P(y_i = l_2 | x_i) > t_{class}$$

Here $l_1$ is the most likely class label and $l_2$ is the second most likely class label, according to the MLM. This is motivated by the \textit{Breaking Ties} active learning method of \citet{Scheffer2001ActiveHM, JMLR:v6:luo05a}, which aims to select token samples where the difference between the top two predictions is the smallest, in order to increase the likelihood of confident classifications. However, for the MLM, we adopt the \textit{inverse} of breaking ties, where we maximize the difference between the top two predictions, based on a threshold. We use different thresholds for each class as shown in Table \ref{tab:mlm-thresholds}. We empirically observe that we obtain a slightly higher F1 score with the MLM as an unsupervised NER on the CoNLL-03 development set when using different thresholds for each class instead of a single threshold value for all classes.

In our experiments, we also empirically observe that the MLM tends to produce more robust probabilities when the lexicon entities filling the {\tt [MASK]} slot(s) are segmented into fewer subwords by the model's tokenizer. This is supported by the findings of \citet{Kauf2023ABW}, who observe that methods that estimate the psuedo-log-likelihood of a sentence yield inflated scores for out-of-vocabulary
words. Hence we employ an additional heuristic where we filter the lexicon entities to only single subword entities. We believe that better methods for estimating and aggregating probabilities for sentences that contain out-of-vocabulary words can be explored in future work. \citet{Kauf2023ABW} introduce one such method, which has been shown to address the issue of attributing uneven likelihoods to multi-token words. Specifically, it proves beneficial to mask not only the current token but also all subsequent tokens that are part of the same word.

Furthermore, given the large size of the lexicons extracted for the MLM in both the ``5\%'' and fully supervised setting, we filter our lexicon to keep only the top 20 entities for each class, sorted by their frequency of occurrence in the training data.

\begin{table}[h]
\centering
\begin{tabular}{cc}
\toprule
Class & Threshold \\
\midrule
ORG & 0.28 \\
PER & 0.2 \\
LOC & 0.1 \\
MISC & 0.05 \\
\bottomrule
\end{tabular}
\caption{\footnotesize{Per class thresholds used by the Masked Language Model (MLM) to implement ``inverse breaking ties''.}}
\label{tab:mlm-thresholds}
\end{table}

\section{Hyperparameters and Hardware}
\label{sec:appendix4}
Instead of the regular cross-entropy loss, we use a Generalized Cross Entropy Loss function with label smoothing \cite{Dimachkie_2023} for training our models, which offers a better trade-off between the noise-robustness of mean absolute error and the noise sensitivity of cross entropy loss. This trade-off can be controlled by a hyperparameter $q$. We experiment with multiple settings where we vary $q$, along with the learning rate, the dynamic window size, the number of burn-in, intermediate and burn-out stages and the total number of self-training iterations. This search involved under 20 runs, based on the development partition of CoNLL-2003. We use a dynamic window size of $5$, a batch size of $16$ for training, a learning rate of $1e{-5}$, and a confidence-threshold of $0.9$ across all of our data settings. We show the other hyperparameters in Table \ref{table:hyperparams}. All experiments were carried out on a system with 2 Nvidia RTX 3090 GPUs.

\begin{table*}[p] 
\centering
\small
\renewcommand{\arraystretch}{3.0} 
\setlength{\tabcolsep}{4pt} 
\setlength{\heavyrulewidth}{1.5pt} 
\setlength{\lightrulewidth}{1.0pt} 
\newcolumntype{Y}{>{\centering\arraybackslash}X}

\begin{tabularx}{\textwidth}{@{}YYYYYYY@{}}
\toprule
\textbf{Setting} & \textbf{Burn-in stages} & \textbf{Intermediate stages} & \textbf{Burn-out stages} & \textbf{Noise-level (\texttt{q})} & \textbf{Self-training iterations} & \textbf{Label Smoothing} \\
\midrule
1\% data & 1 & 2 & 1 & 0.9 & 4 & 0.1 \\
5\% data & 1 & 2 & 0 & 0.7 & 3 & 0.1 \\ 
100\% data & 1 & 1 & 0 & 0.7 & 2 & 0.2 \\
\bottomrule
\end{tabularx}
\caption{The Hyperparameters we use for training ELLEN under various supervision settings.}
\label{table:hyperparams}
\end{table*}

\end{document}